\documentclass{article}

 \usepackage[preprint]{neurips_2026}

\usepackage[utf8]{inputenc} 
\usepackage[T1]{fontenc}    
\usepackage{hyperref}       
\usepackage{url}            
\usepackage{booktabs}       
\usepackage{amsfonts}       
\usepackage{nicefrac}       
\usepackage{microtype}      
\usepackage{xcolor}         
\usepackage{xspace}
\usepackage{graphicx}
\usepackage{enumitem}
\usepackage{amsmath}
\usepackage{cleveref}
\usepackage{subcaption}
\usepackage{wrapfig}
\usepackage{graphicx}
\usepackage[table]{xcolor}
\newcommand{\benchmark}{\textsc{SE-Bench}\xspace}
\title{\textsc{SE-Bench}: Benchmarking Self-Evolution with Knowledge Internalization}
\usepackage[most]{tcolorbox}

\newtcolorbox{takeaway}{
    colback=gray!5,     
    colframe=black!75,  
    width=\columnwidth, 
    arc=1mm,            
    boxrule=0.5pt,      
    left=5pt, right=5pt, top=5pt, bottom=5pt, 
    fonttitle=\bfseries,
    title=Takeaway,     
    after skip=10pt,    
    before skip=10pt    
}

\author{%
  Jiarui Yuan\footnotemark[1], Tailin Jin\footnotemark[1], Weize Chen\footnotemark[1], Zeyuan Liu\\
  Tsinghua University\\
  \texttt{\{yuanjr22,jintl23,chenwz21\}@mails.tsinghua.edu.cn}
}

\begin{document}

\maketitle

\begin{abstract}
  True self-evolution requires agents to act as lifelong learners that \textbf{internalize} novel experiences to solve future problems. However, rigorously measuring this foundational capability is hindered by two obstacles: the \textit{entanglement of prior knowledge}, where ``new'' knowledge may appear in pre-training data, and the \textit{entanglement of reasoning complexity}, where failures may stem from problem difficulty rather than an inability to recall learned knowledge. We introduce \benchmark, a diagnostic environment that obfuscates the NumPy library and its API doc into a pseudo-novel package with randomized identifiers. Agents are trained to internalize this package and evaluated on simple coding tasks without access to documentation, yielding a clean setting where tasks are trivial with the new API doc but impossible for base models without it. Our investigation reveals three insights: (1) the \textbf{Open-Book Paradox}, where training with reference documentation inhibits retention, requiring "Closed-Book Training" to force knowledge compression into weights; (2) the \textbf{RL Gap}, where standard RL fails to internalize new knowledge completely due to PPO clipping and negative gradients; and (3) the \textbf{viability of Self-Play} for internalization, proving models can learn from self-generated, noisy tasks when coupled with SFT, but not RL. Overall, \benchmark establishes a rigorous diagnostic platform for self-evolution with knowledge internalization. Our code and dataset can be found at https://github.com/thunlp/SE-Bench
\end{abstract}

\section{Introduction}
\label{sec:introduction}
Self-evolution, the capacity for an autonomous agent to recursively improve its own capabilities, is often viewed as a prerequisite for Artificial General Intelligence (AGI)~\citep{DBLP:series/cogtech/354023733,DBLP:conf/agi/LeggH06}.
An ideal self-evolving agent acts as a \textit{lifelong} learner, continuously assimilating information from its environment, optimizing its solutions, and expanding its skill set without human intervention.
However, current approaches often limit the scope of this evolution to transient or localized adaptations, such as inference-time response refinement~\cite{DBLP:journals/corr/abs-2506-13131,DBLP:journals/corr/abs-2511-23473} or iterative self-code modification~\cite{DBLP:journals/corr/abs-2505-22954,DBLP:journals/corr/abs-2510-21614}. While valuable, these mechanisms differ fundamentally from the expansive definition we explore here: the self-evolution requires agents to actively learn from experience by \textit{internalizing} novel skills or knowledge, akin to a human expert accumulating domain knowledge over time~\citep{DBLP:journals/tmlr/WangX0MXZFA24,DBLP:journals/corr/abs-2510-04618,DBLP:journals/corr/abs-2509-25140}.

Despite rapid progress in large language model (LLM) reasoning capabilities, we lack a rigorous measurement for this foundational \textit{internalization} ability.
Existing benchmarks have made strides in evaluating specific sub-skills related to self-evolution, such as long-horizon information retrieval~\citep{DBLP:journals/corr/abs-2504-12516,DBLP:journals/corr/abs-2508-13186}, iterative response refinement~\citep{DBLP:journals/corr/abs-2511-22173}, and complex task execution~\citep{tbench_2025,DBLP:conf/iclr/JimenezYWYPPN24}
However, current evaluations fail to cleanly isolate an agent's ability to process and restore experience due to two fundamental obstacles.
First, \textbf{the entanglement of prior knowledge}: when a model solves a task involving ``novel'' knowledge, it is indistinguishable whether the agent learned from the relevant experience or merely recalled pre-training data.
Second, \textbf{the entanglement of reasoning complexity}: if an agent fails a complex task, it is ambiguous whether it failed to \textit{internalize} the necessary knowledge or failed to \textit{reason} over it.
This mirrors a student who memorizes a textbook but fails a frontier math problem due to logical difficulty rather than memory gaps.

To address these limitations, we argue that the community needs a ``Needle in a Haystack'' test~\citep{needle-in-haystack} for self-evolution: an environment where tasks are algorithmically \textit{trivial} if knowledge is internalized, and \textit{impossible} if it is not.
To this end, we introduce \benchmark. \benchmark relies on a \textit{knowledge obfuscation} mechanism to create this clean environment.
We employ a \textit{knowledge obfuscation} mechanism, mapping the core functions of the NumPy~\citep{harris2020array} library to randomized, nonsense identifiers (e.g., \texttt{numpy.mean} $\rightarrow$ \texttt{zwc.kocito}) and rewriting the documentation to describe a ``new'' package.
At training time, agents have access to the documentation, but at test time, agents are tasked with solving simple problems using this obfuscated package, with the strict constraint that any use of the original NumPy library is deemed to fail.
This design grants \benchmark three diagnostic properties:
\textbf{(1) Impossible without information}: Without documentation, the probability of guessing the correct API is mathematically zero, eliminating prior knowledge confounds.
\textbf{(2) Trivial with information}: Because the underlying logic maps 1-to-1 to standard NumPy, tasks are trivial for any agent that internalizes the mapping. Any ideal self-evolving method should theoretically achieve a near-100\% success rate, thus cleanly isolating the internalization capability.
\textbf{(3) Compositional generalization}: While the training set consists of tasks solvable with single function calls, the test set requires composing multiple internalized functions, assessing generalization beyond simple memorization.


Beyond serving as a rigorous metric, \benchmark functions as a clean testbed that enables us to dissect the fundamental mechanisms of self-evolution. We investigate whether standard parameter-optimization paradigms, specifically Supervised Fine-Tuning (SFT) and Reinforcement Learning (RL), can genuinely support the \textit{internalization} capability. 
Our experiments uncover three critical insights: (1) \textbf{The Open-Book Paradox:} We find that the presence of reference material during parameter update inhibits long-term retention. True internalization requires \textit{Closed-Book Training}: removing the documentation during parameter updates forces the model to compress external logic into its weights, significantly outperforming standard SFT. (2) \textbf{The RL Gap}: While SFT effectively internalizes new knowledge, standard RL fails even under the Closed-Book training setting. We identify that the negative gradient and PPO clipping~\citep{DBLP:journals/corr/SchulmanWDRK17} both are factors that impact the knowledge internalization for RL. (3) \textbf{Viability of Self-Play:} By applying SFT instead of RL to self-generated tasks and corresponding responses, models successfully internalize knowledge from their own noisy, unverified data, proving that self-evolution on knowledge internalization is viable if the correct optimization mechanism is used and that RL is not a one-size-fits-all solution.

We position \benchmark as a diagnostic testbed for the self-evolving agent community.
Just as long-context models must at least demonstrate near-perfect retrieval on Needle-in-a-Haystack tests to establish basic competency, we argue that self-evolving agents should also demonstrate the ability to pass \benchmark before they can be trusted to evolve in complex, open-ended environments.
And because \benchmark provides a clean, controlled environment, it also serves as an ideal platform for studying the fundamental mechanisms for knowledge internalization, potentially facilitating future research.


\vspace{-0.7em}
\section{\benchmark}
\label{sec:method}

\begin{figure*}[t!]
    \centering
    \includegraphics[width=\linewidth]{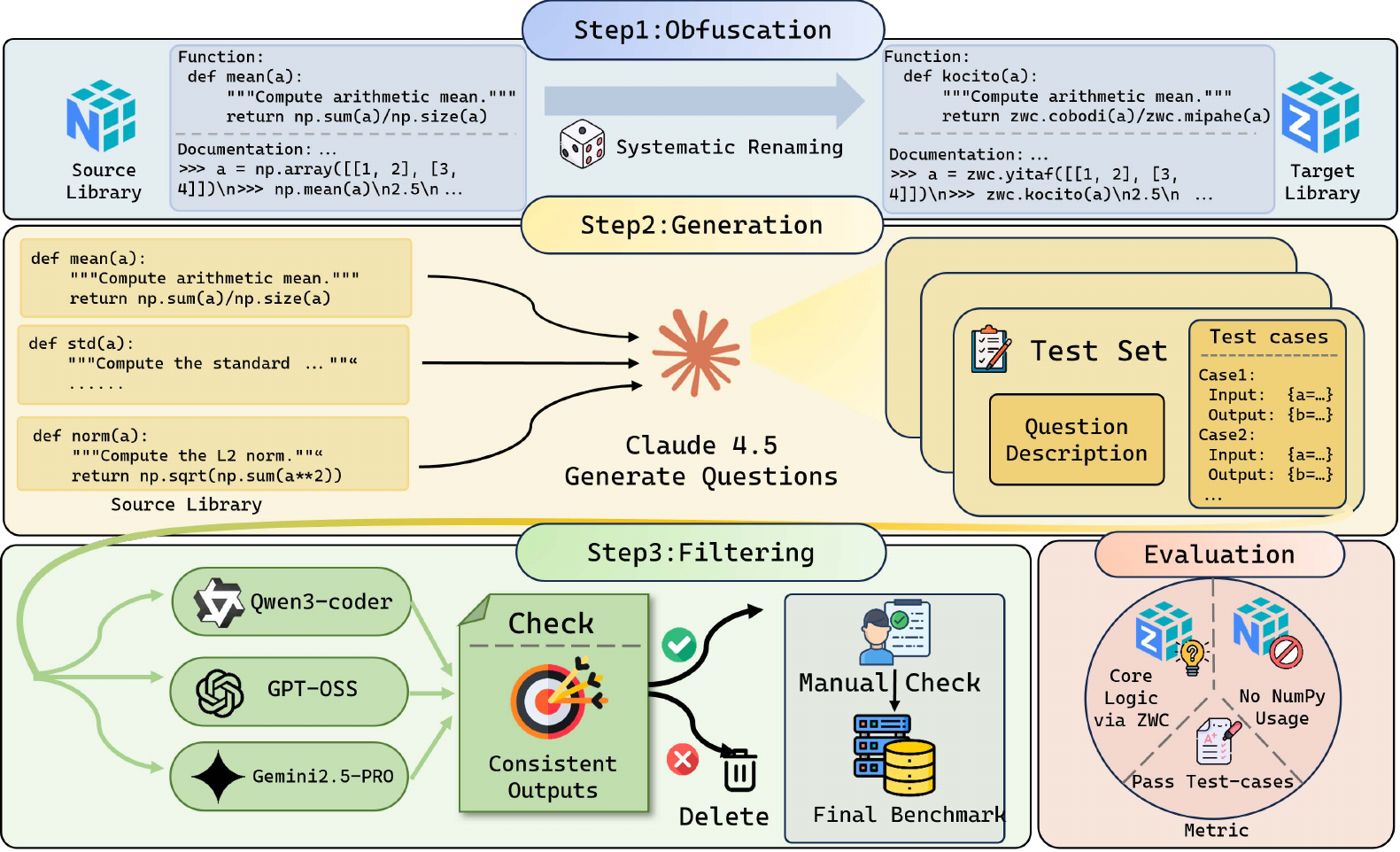}
    \vspace{-0.7em}
    \caption{\textbf{Overview of the \benchmark construction pipeline.} The process consists of three main stages: \textbf{(1) Obfuscation}, where we implement a wrapper package \texttt{zwc} that renames selected NumPy functions and translates API documentation; \textbf{(2) Generation}, where Claude-4.5-sonnet generates valid tasks and test cases based on the original NumPy library; and \textbf{(3) Filtering}, where tasks are validated through strict consensus between three strong LLMs, followed by human verification.}
    \label{fig:benchmark}
    \vspace{-1.3em}
\end{figure*}

\vspace{-0.3em}
We argue that a fundamental, yet often overlooked, component of self-evolution is \textit{knowledge internalization}. While current methods often focus on \textit{transient} adaptation, optimizing a solution within a single context window~\cite{DBLP:journals/corr/abs-2511-23473,DBLP:conf/iclr/QiLILSSYYY00D25,DBLP:journals/corr/abs-2502-18449,qwen3technicalreport}, genuine evolution requires transitioning from a stateless processor to a \textit{lifelong learner}. A concrete example of such a process is a human software engineer learning a new library: initially relying on documentation, but eventually internalizing the logic to solve problems fluently without external aid through repeated practice.

Measuring such a capability of LLM agents in a similar scenario, however, presents a fundamental dilemma. We cannot evaluate the agent's ability to internalize any existing library (like Numpy), as they may have already been embedeed in the LLM's pre-training weights~\cite{DBLP:journals/corr/abs-2506-10947,DBLP:journals/corr/abs-2507-10532}. Furthermore, simply adopting a newly-released library is a fragile solution: as knowledge cutoff of the LLMs advances, the benchmark quickly becomes obsolete. To rigorously measure the \textit{internalization} ability of self-evolving methods, we require a domain that is \textit{permanently} out-of-distribution: one that effectively does not exist on the Internet, ensuring that no further model can solve it with its pre-training knowledge.
 
To this end, we introduce \benchmark, a synthetic domain constructed by systematically obfuscating the NumPy library paired with trivial coding problems. By mapping function names to nonsense identifiers while preserving the underlying logic, we create a ``novel'' package that remains structurally realistic yet alien to any model's training distribution. This construction enforces three critical properties:
\begin{itemize}[leftmargin=*, noitemsep, topsep=-0.2em]
    \item \textbf{Impossible without Information:} The randomized namespace guarantees a mathematical zero-shot baseline, eliminating pre-training confounds.
    \item \textbf{Trivial with Information:} Because the logic is isomorphic to standard NumPy, tasks are algorithmically trivial if the new API doc is provided, cleanly isolating \textit{memory} failures from \textit{reasoning} failures.
    \item \textbf{Compositional Generalization:} By retaining the library's original structure, we can evaluate whether agents can compose internalized functions to solve multi-step problems beyond their specific training examples that involves only a single function.
\end{itemize}

\vspace{-0.2em}
\subsection{Benchmark Construction}

\vspace{-0.2em}
To ensure that \benchmark serves as a rigorous evaluation of internalization, we implement a three-stage construction pipeline: \textit{Obfuscation}, \textit{Question Generation}, and \textit{Filtering}. This process is illustrated in \cref{fig:benchmark}.


\textbf{Stage I: Obfuscation.}
We select NumPy as our source domain due to its functional richness and simplicity. To construct our target library \textbf{ZWC} (a randomly generated package name), we identify a set of 268 common NumPy functions (see \cref{appendix:numpy_functions}) to serve as the core of the new package.
Rather than simply renaming functions, we implement ZWC as a wrapper package. Each function in ZWC is assigned a randomized, semantically void identifier (\textit{e.g.}, \texttt{zwc.kocito}) which internally calls the corresponding NumPy function.
And to prevent models from bypassing the obfuscation by invoking standard methods on the returned NumPy array objects (e.g., calling \texttt{.mean()} directly on an array), we wrap all inputs and outputs in a custom \texttt{ZWCArray} class. This ensures that the agent must strictly rely on the obfuscated functional API to manipulate data.
To rewrite the accompanying documentation, we employ Gemini-2.5-Pro~\citep{DBLP:journals/corr/abs-2507-06261}. We provide the model with the original NumPy docstrings and the global function mapping, instructing it to translate the documentation into the context of the new ZWC package. This results in an API documentation that describes the ``new'' package.

\textbf{Step II: Question Generation.}
Prompting a model to generate tasks directly using the obfuscated \textbf{ZWC} library is prone to hallucination, as the model lacks prior exposure to the new syntax.
We therefore prompt Claude-4.5-sonnet~\citep{anthropic2025claude-sonnet-4-5} to generate simple coding problems relevant to the sampled \textit{original} NumPy functions along with at least 8 test-cases. We employ a stratified generation strategy to create two distinct task categories:
\begin{itemize}[leftmargin=*, noitemsep, topsep=-0.5em]
    \item \textbf{Single-Function Tasks:} We iterate through every function in our function list. For each function, we prompt the model to create a self-contained problem that requires specifically that function to solve. This ensures 100\% coverage of the functions included in \benchmark.
    \item \textbf{Multi-Function Tasks:} To test generalization, we randomly sample sets of 10 functions and prompt the model to generate a complex problem that requires the composition of at least three functions from the sampled set.
\end{itemize}


\textbf{Step III: Filtering.}
To ensure the \textit{Trivial with Information} property, we must verify that the generated tasks are not accidentally difficult or erroneous. We employ a strict \textit{Consensus Filtering} protocol.
We provide the generated questions (in their original NumPy form) to three distinct state-of-the-art models: Qwen3-Coder-480B~\citep{qwen3technicalreport}, Gemini-2.5-Pro~\citep{DBLP:journals/corr/abs-2507-06261}, and GPT-OSS-120B~\citep{openai2025gptoss120bgptoss20bmodel}.
A task is retained only if \textit{all three} models can independently solve it and pass all test cases using standard NumPy. If three distinct model families can solve the question easily, we can be confident that a failure of a testing model on \benchmark is due to a failure to learn the new API, not any error in the problem itself.
Finally, we perform human verification on a 10\% random subset of the filtered data to ensure the clarity of the problem descriptions; all sampled tasks were found to be valid.

\vspace{-0.5em}
\subsection{Dataset Splits \& Protocol}
We partition the filtered tasks into training and testing sets to rigorously evaluate the knowledge internalization.
For \textbf{training set}, it only includes \textbf{Single-Function} tasks, and ensures that every function in the \textbf{ZWC} library appears at least once.
For \textbf{test set}, it comprises both \textbf{Single-Function} tasks (to test retention on unseen problems) and \textbf{Multi-Function} tasks (to test compositional generalization).

The core objective of \benchmark is to measure internalization. Therefore, the information availability differs strictly between phases:
\begin{itemize}[leftmargin=*, noitemsep, topsep=-0.5em]
    \item \textbf{Training Phase:} The agent is provided with the training set, each task includes a problem description and the \textit{relevant documentation} for the involved function. The agent may use this phase to practice, memorize, or update its parameters.
    \item \textbf{Testing Phase:} The agent is provided with the test set that includes \textit{only} the problem descriptions without API documentation. To solve the task, the agent must rely entirely on the knowledge internalized during training.
\end{itemize}

\subsection{Metrics}

To ensure rigorous evaluation, we employ a strict Abstract Syntax Tree (AST) verification protocol. A solution is not judged merely on output correctness, but on its adherence to the benchmark constraints. In the implementation, we further incorporate special-case checks to filter out several rare but challenging hacking behaviors.

Since ZWCArray is convertible to NumPy arrays, models might attempt to bypass the task by converting data to NumPy, performing operations, and converting back. To prevent this, we explicitly prohibit the import or usage of the original \texttt{Numpy} package.

A solution is considered correct ($R(s)=1$) if and only if it meets three conditions: (1) it passes all provided test cases, (2) AST analysis confirms that the returned value relies on ZWC APIs, and (3) it contains zero imports of NumPy:
\begin{equation}
\begin{aligned}
R(s) =
\begin{cases}
1, & \text{if } \text{all 3 conditions are met}, \\
0, & \text{otherwise}.
\end{cases}
\end{aligned}
\label{eq:evaluation}
\end{equation}

\subsection{Statistics and Validation}
\label{sec:statistics}

\benchmark comprises 1,417 tasks, partitioned into a training set of 718 instances and a test set of 699 instances. To evaluate generalization capabilities, we stratify the test set by compositional complexity: it contains 259 \textit{Single-Function} tasks, which require only single API calls, and 440 \textit{Multi-Function} tasks that demand the composition of multiple APIs. Both splits maintain comprehensive coverage of the ZWC surface area, and an illustrative example is provided in \cref{appendix:benchmark_example}.

\begin{wraptable}{r}{0.45\linewidth}
    \centering
    \vspace{-0.8em}
    \small 
    \caption{\textbf{Validation of \benchmark Design Properties.} Pass@64 of Qwen3-8B. \textbf{Standard NumPy} confirms the tasks are reasoning-trivial. \textbf{ZWC Zero-Shot} confirms no pre-training leakage. \textbf{ZWC In-Context} establishes a solvability ceiling when documentation is provided without training.}
    \vspace{0.2em}
    \resizebox{\linewidth}{!}{
    \begin{tabular}{lcc}
        \toprule
        \textbf{Evaluation Setting} & \textbf{Single} & \textbf{Multiple} \\
        \midrule
        \multicolumn{3}{l}{\textit{Reasoning Upper Bound (Using Standard NumPy)}} \\
        \textbf{Standard NumPy} & 97.4 & 93.6 \\
        \midrule
        \multicolumn{3}{l}{\benchmark{} \textit{Strict Evaluation (Using ZWC)}} \\
        \textbf{ZWC Zero-Shot} & 0.0 & 0.0 \\
        \textbf{ZWC In-Context} & 85.3 & 70.5 \\
        \bottomrule
    \end{tabular}
    }
    \label{tab:pass_k}
    \vspace{-0.9em}
\end{wraptable}

To validate the structural integrity of our design, we conducted a preliminary evaluation on our test set using Qwen3-8B under three distinct settings, as reported in \cref{tab:pass_k}. First, we evaluate the \textbf{Standard NumPy} setting, where the the model is allowed to use the original NumPy library. The high accuracy ($>$ 90\%) across both splits confirms that the tasks are reasoning-trivial. Second, we test the \textbf{ZWC Zero-Shot} setting, where the model must use the obfuscated library and is denied access to API documentation. The resulting 0\% accuracy confirms that our obfuscation is robust and prevents any leakage of pre-training knowledge. Finally, we examine the \textbf{ZWC In-Context} setting, where the model is provided with the API documentation relevant to the problem. The recovery of performance demonstrates that the benchmark is solvable. Notably, the remaining gap (compared to $>$90\% on NumPy) is primarily due to hallucination: qualitative analysis reveals that Qwen3-8B frequently tries to use NumPy namespace (\textit{e.g., } \texttt{zwc.\underline{mean}}) despite the provided relevant documentation. This underscores a core objective for self-evolution: methods must enable agents to strictly adhere to real, internalized knowledge.


\vspace{-0.6em}
\section{Experiment}
\label{sec:experiments}
\begin{table*}[t]
    \centering
    \caption{\textbf{Average performance over 5 rollouts of different models and methods}. The best results are highlighted in \textbf{bold}, and the second-best results are \underline{underlined}.}
    \resizebox{\linewidth}{!}{
    \begin{tabular}{l
        r@{\hspace{12pt}}r@{\hspace{12pt}}r@{\hspace{24pt}}
        r@{\hspace{12pt}}r@{\hspace{12pt}}r
    }
    \toprule
      & \multicolumn{3}{c}{\textbf{Single}} & \multicolumn{3}{c}{\textbf{Multiple}} \\
    \cmidrule(lr){2-4} \cmidrule(lr){5-7}
     \textbf{Method}
     & \textbf{Qwen3-8B} & \textbf{Qwen3-4B} & \textbf{Qwen3-1.7B}
     & \textbf{Qwen3-8B} & \textbf{Qwen3-4B} & \textbf{Qwen3-1.7B} \\
    \midrule

    \textit{Memory Based} \\ \textbf{ACE} \citep{DBLP:journals/corr/abs-2510-04618}   & 11.2 & 12.0 & 2.0 & 4.1 & 6.6 & 0.0 \\
                          \textbf{Expel} \citep{DBLP:conf/aaai/Zhao0XLLH24}   & \underline{47.1} & \underline{39.4} & \textbf{28.2} & \underline{15.5} & \textbf{13.6} & \textbf{4.8} \\
    \midrule

    \textit{SFT Based}    \\ 
                          \textbf{Open-SFT}        & 0.0     & 0.0 & 0.0 & 0.0     & 0.0 & 0.0 \\
                          \textbf{Closed-SFT}     & 39.6 & 25.1 & 16.9 & 11.6 & 5.4 & 2.0 \\
    \midrule

    \textit{RL Based}     \\
                          \textbf{Open-RL} & 0.0     & 0.0 & 0.0 & 0.0     & 0.0 & 0.0 \\
                          \textbf{Closed-RL} & 0.0 & 0.0 & 0.0 & 0.0 & 0.0 & 0.0 \\
                          \textbf{Absolute-Zero} \citep{DBLP:journals/corr/abs-2505-03335} & 0.0 & 0.0 & 0.0 & 0.0 & 0.0 & 0.0 \\
    \midrule
    \textit{Hybrid}      \\
                          \textbf{Closed-SFT-RL} & \textbf{54.4} & \textbf{43.1} & \underline{21.9} & \textbf{17.9} & \underline{9.2} & \underline{3.2} \\
    
    \bottomrule
    \end{tabular}
    }
    \vspace{-1.8em}
    \label{tab:main-table}
\end{table*}

\textbf{Baselines: }
We evaluate a diverse suite of self-evolution strategies across three paradigms. \textbf{(1) Memory-based:} \textbf{ACE}~\citep{DBLP:journals/corr/abs-2510-04618} and \textbf{Expel}~\citep{DBLP:conf/aaai/Zhao0XLLH24}, which summarize experience into memory to improve future task performance. \textbf{(2) Parameter-Optimization (SFT/RL):} We consider two fundamental training protocols. In the \textbf{Open} setting, API documentation remains in the context during both trajectory collection and parameter updates. In the \textbf{Closed} setting, documentation is available for trajectory collection but is \textit{stripped} during training. See \cref{appendix:difference_between_open_and_closed} for the difference between the Open and Closed settings. We also evaluate the fully autonomous self-play method \textbf{Absolute-Zero}~\citep{DBLP:journals/corr/abs-2505-03335}. \textbf{(3) Hybrid:} \textbf{Closed-SFT-RL} applies standard RL without API documentation on top of the Closed-SFT.


\textbf{Training Setup: } We use Qwen3-8B, 4B and 1.7B ~\cite{qwen3technicalreport} families as base models. All RL-based methods utilize the GRPO algorithm~\citep{DBLP:journals/corr/abs-2402-03300} within the veRL~\cite{sheng2024hybridflow} framework. Experiments were conducted on 8 NVIDIA A100 GPUs; further details are provided in \cref{appendix:main_experiment_details}. To validate that our findings generalize to larger-scale models and different model families, we also conduct supplementary experiments on Qwen3-30B-A3B and Llama3.2-3B. Detailed results are provided in \cref{appendix:additional_experiments}.

\textbf{Results.}
\cref{tab:main-table} showcases baseline performance on \benchmark. Memory-based methods achieve non-trivial results across all model sizes, with \textbf{Expel} attaining the highest accuracy on the Multi-Function split. This is intuitive, as memory allows the model to map \benchmark identifiers to NumPy correspondents, and then revise the NumPy solution based on the mapping. However, even these methods remain far from perfect, indicating that autonomous memory management is still in its infancy.


Even more surprisingly, among parameter-update methods, only \textbf{Closed-SFT} and the hybrid \textbf{Closed-SFT-RL} achieve success; all other methods fail completely. This striking gap suggests that standard RL is fundamentally ill-suited for knowledge internalization, a failure we analyze mechanistically in \cref{sec:RL_internalize_knowledge_analysis}. Furthermore, the comparison between \textbf{Open-SFT} and \textbf{Closed-SFT} reveals the \textbf{Open-Book Paradox}: despite using the same training trajectories, success depends entirely on removing documentation during parameter updates. This forces the model to encode logic into its weights rather than relying on context. As we will show in \cref{sec:sft_internalize_knowledge_analysis}, this reflects genuine internalization rather than simple alignment to the test-time prompt distribution.
\vspace{-0.6em}
\section{Analysis and Insight}
\label{sec:analysis}
    
    
\begin{wrapfigure}{r}{0.5\linewidth}
    \centering
    \vspace{-3.0em}
    \begin{subfigure}[t]{0.49\linewidth}
      \centering
      \includegraphics[width=\linewidth]{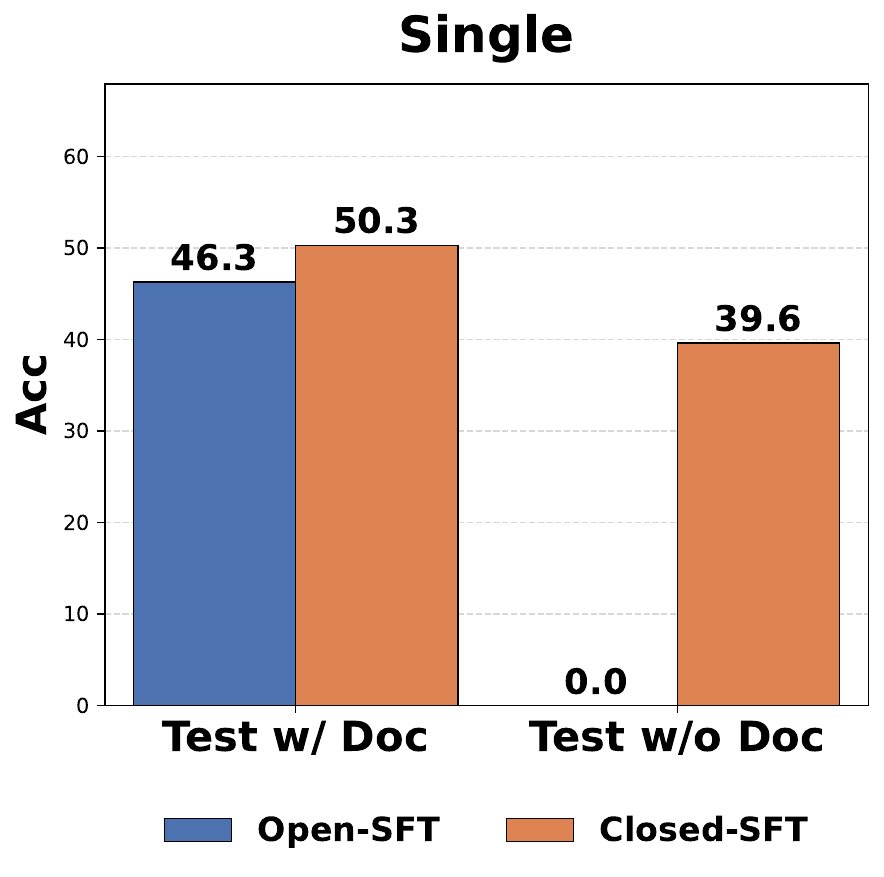}
      \vspace{-1.5em}
      \label{fig:easy_prov_doc_abla}
    \end{subfigure}%
    \hfill
    \begin{subfigure}[t]{0.49\linewidth}
      \centering
      \includegraphics[width=\linewidth]{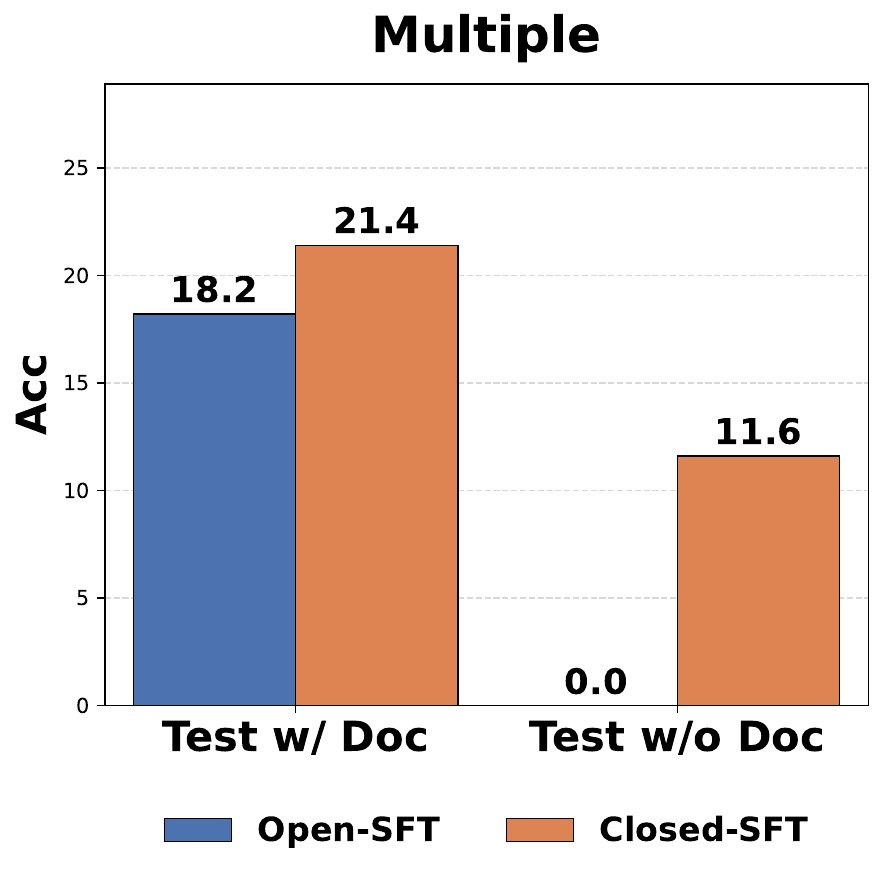}
      \vspace{-1.2em}
      \label{fig:hard_prov_doc_abla}
    \end{subfigure}
    
    \caption{\textbf{Closed-SFT vs. Open-SFT}. Open-SFT's complete failure without documentation reveals strict context dependency, whereas Closed-SFT successfully internalizes knowledge, maintaining performance even when documentation is absent.}
    \label{fig:sft_internalize}
    \vspace{-2.0em}
\end{wrapfigure}
Beyond serving as a rigorous metric for current methods, \benchmark functions as a clean testbed for investigating the fundamental mechanisms of self-evolution with knowledge internalization. 
In this section, we leverage this controlled environment to investigate three fundamental research questions (RQs) about knowledge internalization with parameter update methods: \textbf{(RQ1)}: Does SFT induce true internalization, or merely context dependence? \textbf{(RQ2)}: Can RL internalize knowledge presented in the context? \textbf{(RQ3)}: Can Self-Play Enable Knowledge Internalization? \textbf{(RQ4)}: How does knowledge evolve from SFT to RL?





\subsection{RQ1: Does SFT Induce True Internalization, or Merely Context Dependence?}
\label{sec:sft_internalize_knowledge_analysis}

In \cref{sec:experiments}, we had a surprising observation: removing API documentation during parameter update significantly boosts performance on \benchmark. We hypothesize that this training condition forces the model to rely on parametric memory rather than context. A critical question remains: does this improvement stem from genuine \textit{knowledge internalization}, or is it merely an artifact of distribution consistency between training and testing prompts?

To isolate the mechanism, we evaluate all models with API documentation provided at test time, strictly aligning with the Open-SFT training prompt. As shown in \cref{fig:sft_internalize}, Closed-SFT still outperforms Open-SFT even in this setting. This rules out prompt consistency as the driver; if it were, Open-SFT would lead. Instead, the results confirm that withholding documentation during training forces the model to encode knowledge directly into its parameters, resulting in robust internalization independent of context availability.


\subsection{RQ2: Can RL Internalize Knowledge in the Context?}
\label{sec:RL_internalize_knowledge_analysis}

While Closed-SFT effectively internalizes knowledge, we find that applying RL in the similar off-policy settings, \textit{i.e.,} generating rollout with API doc, and training without it, fails completely. As shown in \cref{tab:main-table}, Closed-RL achieves zero performance. This stark contrast suggests that the mechanism for knowledge internalization is fundamentally different or more restricted in RL than in SFT.

To isolate the cause of this failure, we perform a systematic ablation study bridging the mathematical gap between SFT and RL.
For a prompt $x \sim \mathcal{D}$, let $\{y_i\}_{i=1}^N$ denote $N$ trajectories. The SFT objective maximizes the log-probability of successful trajectories:
\begin{equation}
\mathcal{L}_{\text{SFT}}(\theta)
=
\mathbb{E}_{x \sim \mathcal{D}}
\left[
\frac{1}{N}
\sum_{i=1}^N
\log \pi_\theta(y_i \mid x)
\right].
\label{eq:sft_loss}
\end{equation}

For RL, we omit the KL regularization. And although the off-policy (rollout with doc, train without) theoretically necessitates an importance sampling ratio $\frac{\pi_\theta(y \mid x_{\text{no\_doc}})}{\pi_{\theta}(y \mid x_{\text{doc}})}$, we exclude it in practice as the numerator vanishes for randomized ZWC function names. The training objective is thus formalized as:
\begin{equation}
\mathcal{L}_{\text{RL}}(\theta)
=
\mathbb{E}_{x \sim \mathcal{D}}
\left[
\frac{1}{N}
\sum_{i=1}^N
\rho_i(\theta)A_i
\right].
\label{eq:unified_loss}
\end{equation}
By instantiating the policy term $\rho_i(\theta)$ and the advantage term $A_i$ differently, we recover both paradigms:
\begin{itemize}[leftmargin=*, noitemsep, topsep=0pt]
    \item \textbf{SFT Instantiation:} $\rho_i(\theta) = \log \pi_\theta(y_i \mid x)$ and $A_i = \mathbb{I}(y_i \text{ is correct})$. This treats every correct trajectory as a positive reinforcement signal without regularization.
    \item \textbf{GRPO Instantiation:} $\rho_i(\theta)$ uses the clipped probability ratio $\min(r_t(\theta), \text{clip}(r_t(\theta), 1-\epsilon, 1+\epsilon))$ to constrain updates, and $A_i$ is the group-normalized advantage (containing both positive and negative values).
\end{itemize}
\begin{wraptable}{r}{0.5\linewidth}
    \centering
    \vspace{-0.9em}
    \caption{\textbf{Ablation of RL components.} Internalization collapses when using PPO clip loss or including negative advantage.}
    \begin{tabular}{lcc}
        \toprule
        \textbf{Setting} & \textbf{Single} & \textbf{Multiple} \\
        \midrule
        \textbf{SFT-like Closed-RL} & \textbf{51.0} & \textbf{9.4}  \\
        \rowcolor{gray!15}
        \quad w/o Larger LR   & \underline{31.7} & \underline{1.4}  \\
        \rowcolor{gray!15}
        \quad w/o Larger BSZ   & \underline{31.7} & 0.2 \\
        \rowcolor{gray!15}
        \quad w/ PPO Clip Loss & 0 & 0 \\
        \rowcolor{gray!15}
        \quad w/ GRPO Advantage   & 0 & 0 \\
        \bottomrule
    \end{tabular}
    \label{tab:rl_abla}
    \vspace{-0.6em}
\end{wraptable}
We start by constructing a \textbf{SFT-like Closed-RL} baseline: we configure the RL approach to use SFT-style hyperparameters (High LR, Large Batch), the SFT-style objective ($\rho=\log \pi$), and Binary Advantage ($A \in \{0,1\}$). As shown in \cref{tab:rl_abla}, this configuration successfully recovers Closed-SFT performance, proving that the RL \textit{framework} itself is not the issue.

We then systematically revert each component to its standard GRPO setting to identify the bottleneck. Reverting to standard RL hyperparameters (\textbf{w/o Larger LR/BSZ}), drops performance to 31.7\%, but the model \textit{still learns}, indicating that optimization efficiency is affected rather than fundamental capability. In sharp contrast, the training objective and advantage formulation prove critical. Reintroducing the PPO clipping mechanism (\textbf{w/ PPO Clip Loss}) causes immediate collapse to \textbf{0\%}. Similarly, reintroducing the standard normalized advantage (\textbf{w/ GRPO Advantage}), which introduces negative reinforcement signals, also results in a total collapse to \textbf{0\%}.

This isolates the failure to two specific “safety” mechanisms in standard RL. First, clipping prevents internalization. Internalizing a new vocabulary item (e.g., mapping \texttt{np.mean} to \texttt{zwc.kocito}) requires a substantial shift in probability mass. However, the clipping mechanism penalizes such large updates, effectively suppressing the model’s ability to assign higher probability to newly introduced tokens. As a result, the model is hindered from encoding new definitions.

Second, standard normalized advantage generates negative signals for below-average responses. In the fragile early stages of memorization, these negative gradients likely erase tentative associations before they can solidify. The concrete underlying mechanisms impacting this internalization process remain an interesting direction for future work, and \benchmark provides a clean test-bed for it.

\subsection{RQ3: Can Self-Play Enable Knowledge Internalization?}
\label{sec:self_play_internalization_analysis}
So far, our investigation has relied on carefully curated problems and test cases. A far more compelling scenario is \textit{self-play}: can a model propose its own problems and test cases to internalize knowledge autonomously?

Absolute Zero~\citep{DBLP:journals/corr/abs-2505-03335} is a method that reflect such philosophy, a pure RL loop where the agent self-proposes tasks, and learns to solve. However, as shown in \cref{tab:main-table}, this yields \textbf{0.0\% accuracy}. Given our finding that RL struggles with internalization (RQ2), a critical ambiguity arises: Is this failure due to poor self-generated data (the model cannot teach itself), or simply the improper RL?
To isolate the cause, we investigate whether the model can learn from its own self-proposed curriculum if we switch to SFT. We construct the \textbf{Open-SFT$_\text{self}$} and \textbf{Closed-SFT$_\text{self}$} settings, which are similar to the corresponding settings in~\cref{sec:experiments}, but the questions and test-cases are now generated by the base model itself.

\begin{wraptable}{r}{0.5\linewidth}
    \vspace{-1.0em}
    \centering
    \caption{\textbf{Self-play ablation}. While Absolute Zero fails, applying SFT to the similar autonomous curriculum (\textbf{Closed-SFT$_\text{self}$}) yields significant performance, confirming that self-generated data is sufficient for learning.}
    \resizebox{0.95\linewidth}{!}{
    \begin{tabular}{lcc}
        \toprule
        \textbf{Method} & \textbf{Qwen3-8B (Single)} & \textbf{Qwen3-8B (Multiple)} \\
        \midrule
        \textbf{Absolute Zero} & 0.0 & 0.0 \\
        \textbf{Closed-SFT} & 39.6 & 11.6 \\
        \midrule
        \textbf{Open-SFT$_\text{self}$} & 0.0 & 0.0 \\
        \textbf{Closed-SFT$_\text{self}$} & \textbf{22.5} & \textbf{8.7} \\
        \bottomrule
    \end{tabular}
    }
    \label{tab:selfgen_results}
    \vspace{-1.0em}
\end{wraptable}

The results in \cref{tab:selfgen_results} are decisive. While Absolute-Zero fails completely, \textbf{Closed-SFT$_\text{self}$} recovers significant performance (\textbf{22.5\%}). Although this falls behind the \textbf{Closed-SFT}, which trains on curated problems and test-cases, it demonstrates non-trivial learning. This result explains the failure of standard self-play. The barrier is the \textit{optimization method}, not the self-play paradigm. The model is fully capable of generating valid data conditioned on API documentation to teach itself. The failure occurs strictly because RL can hardly internalize knowledge. By switching the optimization method to SFT, the model now successfully internalizes knowledge \textit{autonomously}.

Besides, consistent with our earlier analysis in \cref{sec:sft_internalize_knowledge_analysis}, \textbf{Closed-SFT$_\text{self}$} outperforms \textbf{Open-SFT$_\text{self}$} regardless of whether we provide documentation during test-time (\cref{fig:sft_self_internalize}), reinforcing the Open-Book Paradox in the self-play setting.
\vspace{-0.7em}
\subsection{RQ4: How Knowledge Evolves from SFT to RL?}
\label{sec:evolution_analysis}


While RQ2 confirms that RL cannot internalize knowledge from scratch, \cref{tab:main-table} shows that the hybrid \textbf{Closed-SFT-RL} achieves state-of-the-art performance among parameter-update methods. This suggests that once SFT internalizes the foundational knowledge, RL acts as a powerful amplifier.

To understand the mechanism of this amplification, we analyze the shift in error patterns from the SFT stage to the RL stage. Specifically, we investigate how RL impact the behavior . We classify the errors into five fine-grained categories (see \cref{appendix:error_type} for examples of each category):
\begin{itemize}[leftmargin=*, noitemsep, topsep=0pt]
    \item \textbf{ZWCArray Attribute Hallucination:} The agent assumes non-existent methods for the \texttt{ZWCArray} object (\textit{e.g.,} guessing \texttt{.tolist()}), reflecting incorrect intuition about the data structure's interface.
    \item \textbf{ZWC Function Hallucination:} The agent invents functions that do not exist in the library (\textit{e.g.,} \texttt{zwc.mean()} instead of \texttt{zwc.kocito()}), indicating a failure to recall the correct API name.
    \item \textbf{Parameter Signature Misalignment:} The agent correctly identifies the function but misremembers its parameter list, leading to execution errors.
    \item \textbf{Return Value Misinterpretation:} The agent misunderstands the output format of a function, causing downstream errors.
    \item \textbf{Native Python Incompatibility:} The agent applies unsupported native Python operations to ZWC objects.
\end{itemize}

\begin{wrapfigure}{r}{0.5\linewidth}
    \centering
    \vspace{-1.6em}
    \begin{subfigure}[t]{0.49\linewidth}
      \centering
      \includegraphics[width=\linewidth]{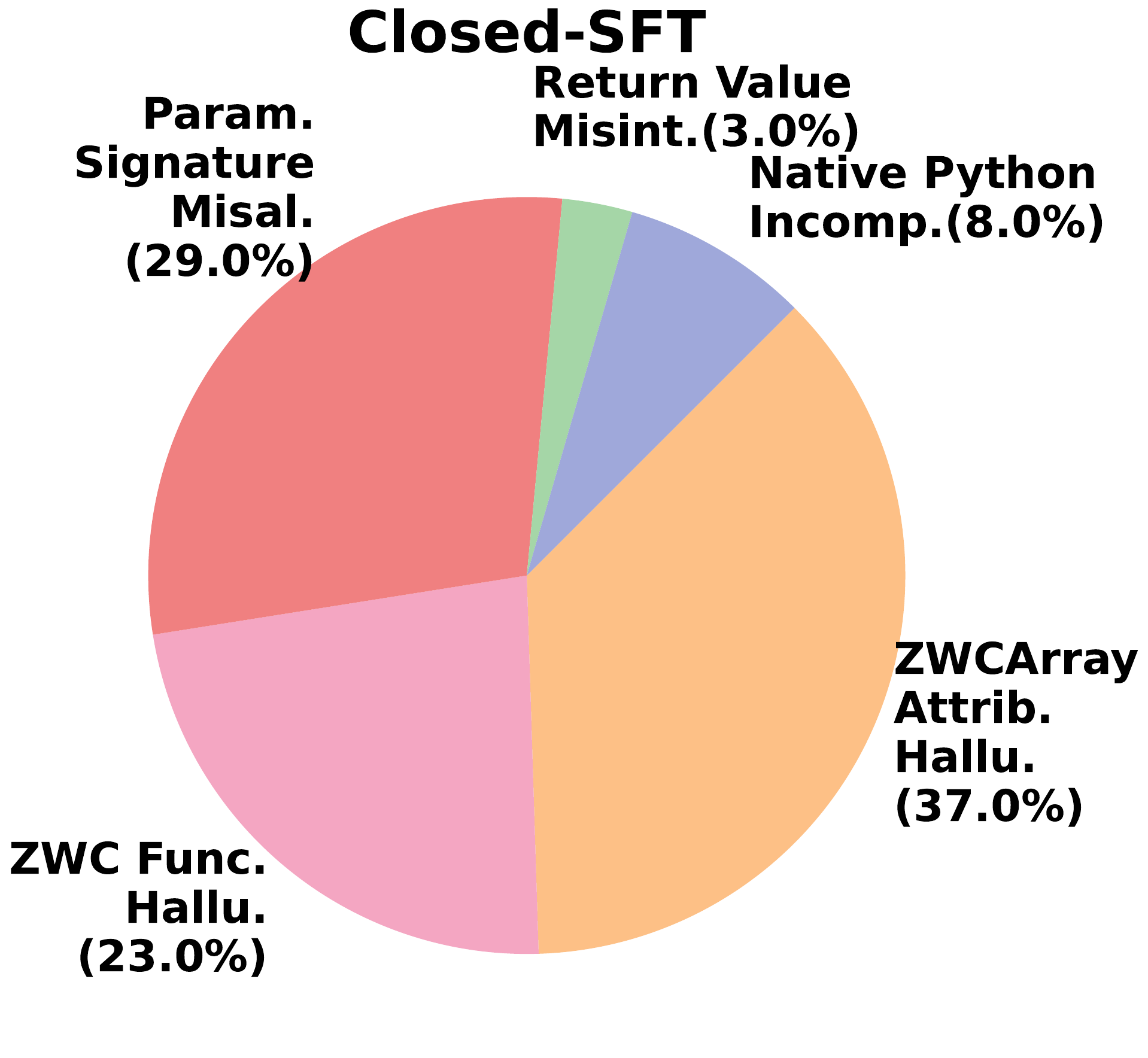}
      \vspace{-4.0em}
      \label{fig:sft-error-type}
    \end{subfigure}%
    \hfill
    \begin{subfigure}[t]{0.49\linewidth}
      \centering
      \includegraphics[width=\linewidth]{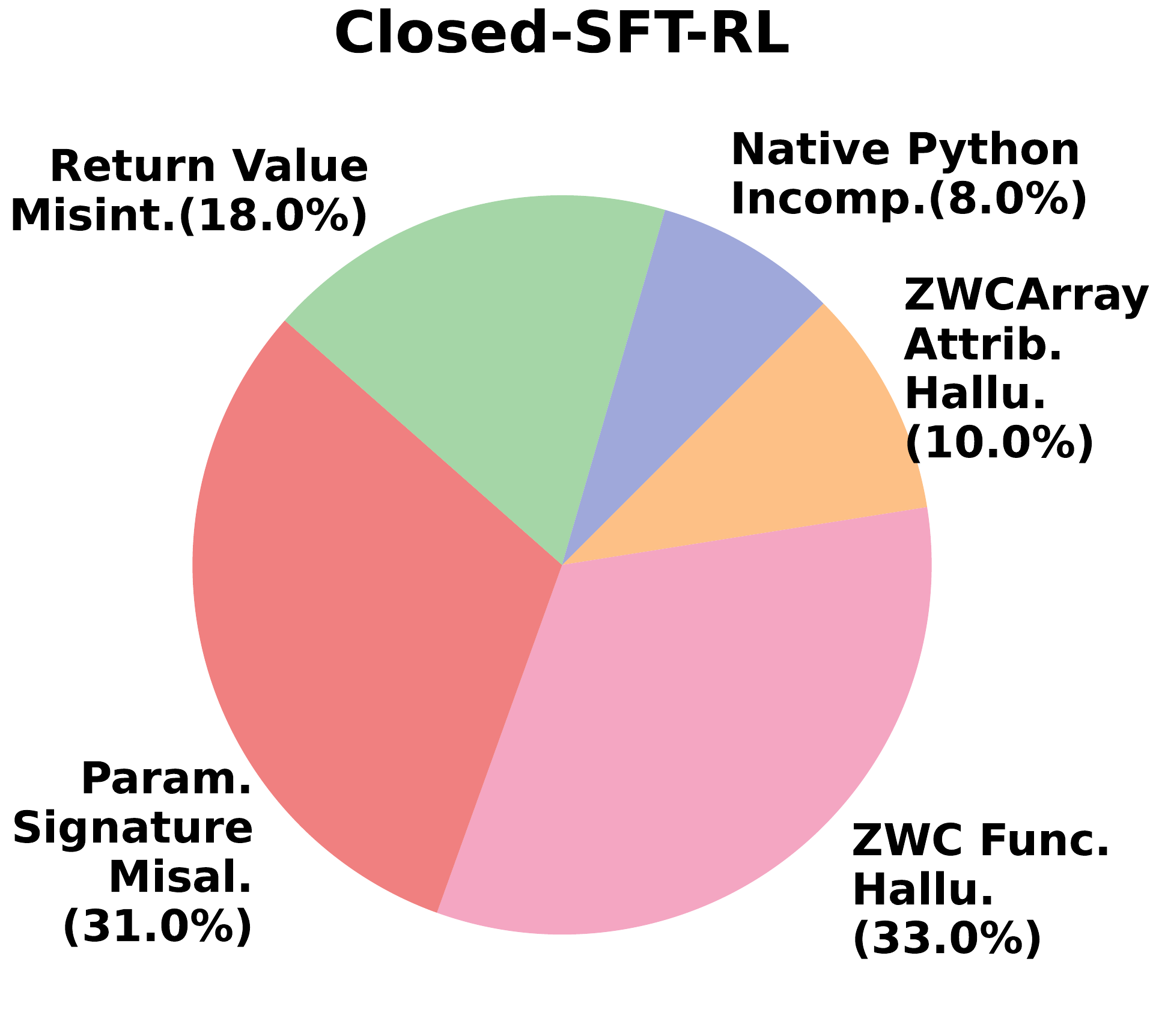}
      \vspace{-1.2em}
      \label{fig:rl-error-type}
    \end{subfigure}
    \caption{Error type distribution for Closed-SFT and Closed-SFT-RL.}
    \label{fig:error-type}
    \vspace{-0.6em}
\end{wrapfigure}

We randomly sampled 100 failed trajectories from both Closed-SFT and Closed-SFT-RL, and used Gemini-3-Flash to classify their error types. \cref{fig:error-type} illustrates the dramatic shift in error distribution.
In the SFT stage, errors are dominated by hallucinations, particularly \textbf{ZWCArray Attribute Hallucination} (37.0\%). This suggests that SFT induces a "probabilistic" form of memory: the model learns the general shape of the library but often fills in gaps with plausible guesses (like assuming ZWCArray has a \texttt{.tolist()} method).

After applying RL, the proportion of \textbf{ZWCArray Attribute Hallucination} collapses to just 10.0\%. Qualitative analysis in our case study (\cref{appendix:case_study}) reveals the mechanism behind this shift: RL does not merely suppress errors; it drives the model to \textit{replace} uncertain API calls with alternative, valid implementations. For instance, when the agent is unsure if a specific ZWC method exists, RL encourages it to fallback to robust primitives (\textit{e.g.,} using explicit loops rather than hallucinated array methods). This results in code that is more disciplined and executable.

However, RL does \textit{not} significantly reduce \textbf{Parameter Signature Misalignment} or \textbf{ZWC Function Hallucination}. This confirms the "RL Gap" identified in RQ2: RL cannot correct fundamental memory errors. If the model mis-memorized a function name or signature during SFT, RL lacks the supervised signal to fix it. Instead, RL optimizes \textit{utilization}, pruning ``lazy" guesses to ensure that what \textit{is} known is applied robustly.

\subsection{Discussion: Connections to Recent Advancements}
\label{sec:discussion}
The mechanisms analyzed in this study, specifically the necessity of \textit{information starvation} for internalization (RQ1) and the distinct roles of SFT and RL (RQ2/3), offer mechanistic insights that complement several recent research directions. \benchmark serves as a controlled environment to isolate and further investigate these dynamics.

For example, our finding that removing documentation during SFT is critical for internalization provides empirical validation for strategies like OpenAI's \textit{Deliberative Alignment}~\citep{DBLP:journals/corr/abs-2412-16339} under knowledge internalization. We quantify the underlying mechanism: removing the relevant information in the context is not merely data cleaning, but a functional requirement that forces the compression of external logic into model parameters.

And recent works in ``Prefix-RL"~\citep{DBLP:journals/corr/abs-2507-01679,qu2026popelearningreasonhard,setlur2026reuseflopsscalingrl} observe that RL trained with privileged prefixes can generalize improvements to un-prefixed settings. our analysis adds crucial nuance. While we confirm RL optimizes knowledge utilization, we highlight its distinct limitation in \textit{internalizing} new factual content compared to SFT. \benchmark enables researchers to rigorously disentangle these two effects, behavioral generalization versus factual internalization, potentially facilitating the design of more targeted hybrid algorithms.
\vspace{-0.6em}
\section{Related Work}
\vspace{-0.4em}
\textbf{LLM powered Agent. }
LLM-powered agents have demonstrated strong effectiveness across a wide range of real-world scenarios. They are capable of performing deep research tasks ~\cite{DBLP:journals/corr/abs-2503-09516, DBLP:journals/corr/abs-2504-03160} and handling code engineering problems ~\cite{DBLP:journals/corr/abs-2401-07339,DBLP:conf/nips/YangJWLYNP24}. However, current approaches for enhancing these agentic capabilities primarily rely on pre-training or post-training with human-annotated data. As a result, continual improvement requires scaling up the amount of labeled data, which is both inefficient and costly, and may eventually encounter a performance ceiling.

\textbf{Self-Evolving Agent. }
To address the limitations above, an agent needs to possess the capability of self-evolution, continuously learning from its own trajectories to improve its performance ~\cite{silver2025welcome}. This ability is widely regarded as a necessary component toward achieving AGI. Methodologically, self-evolution can be realized in several ways. One approach is memory engineering ~\cite{DBLP:conf/aaai/Zhao0XLLH24, DBLP:journals/corr/abs-2510-04618}, where the agent summarizes insights from past trajectories and retrieves them as contextual knowledge when answering similar questions in the future. Another approach is post-training ~\cite{DBLP:journals/corr/abs-2511-23473,DBLP:journals/corr/abs-2508-10874}, which updates the model parameters using previously successful trajectories. In addition, co-evolutionary training methods can be employed to further reduce reliance on real-world annotated data~\cite{DBLP:journals/corr/abs-2505-03335,DBLP:journals/corr/abs-2508-05004}.

\textbf{Evaluating Agents' Evolving Capabilities. }
However, existing evaluations of agent capabilities focus on code generation ~\cite{DBLP:journals/corr/abs-2503-09516,DBLP:conf/iclr/JimenezYWYPPN24}, search~\cite{DBLP:journals/corr/abs-2504-12516}, and tool usage~\cite{DBLP:conf/iclr/MialonF0LS24}. A crucial foundation of self-evolution—the ability to memorize and leverage knowledge—has not been adequately assessed. Current memory benchmarks often focus on learning from user feedback ~\cite{DBLP:journals/corr/abs-2510-17281}, which is weakly verifiable and does not directly reflect improvements in the model’s underlying capabilities. In contrast, our benchmark provides a clean evaluation environment with no data leakage, thereby filling this gap and enabling a more faithful assessment of an agent’s memorize-and-leverage ability.
\section{Conclusion}

We introduce \benchmark, a diagnostic testbed that obfuscates NumPy to test \textit{knowledge internalization}. Our experiments reveal three key insights: (1) the \textbf{Open-Book Paradox}, demonstrating that true retention requires removing knowledge during training, as accessible context inhibits internalization; (2) the \textbf{RL Gap}, showing that standard RL acts as a behavioral optimizer but fails to internalize new facts; and (3) the viability of \textbf{Self-Play}, which, when seeded with SFT, enables models to successfully distill knowledge from their own noisy curricula. We position \benchmark as a critical unit test for future self-evolving agents, ensuring they possess the genuine ability to learn from experience.

{
\small
\bibliographystyle{unsrt}
\bibliography{example_paper}
}

\appendix

\section{Limitation}
\label{appendix:limitaiton}
While \benchmark provides a controlled and effective setting for evaluating an agent’s ability to internalize knowledge during self-evolution, there remain several aspects that need further exploration. 

\textbf{First}, our current design focuses on code generation scenarios, which offer a precise and verifiable testbed for studying internalization. Extending \benchmark to a broader range of task domains—such as those involving factual knowledge—would be a valuable direction to examine the generality of our findings in more diverse and realistic settings.

\textbf{Second}, most of the baselines considered in this work (except for memory-based methods) operate under a single-turn paradigm. As a result, the role of long-horizon, multi-turn interactions in shaping knowledge internalization is not fully explored. Investigating how iterative interaction and feedback influence internalization remains an interesting avenue for future work.

\textbf{Overall}, these considerations point to promising opportunities for extending \benchmark, including broadening task coverage and incorporating richer interaction settings, which may further deepen our understanding of knowledge internalization in self-evolving agents.

\section{The Effect of Question and Trajectory Diversity}
\label{appendix:diversity}
\begin{figure}
    \centering
    
    \begin{subfigure}[t]{0.48\linewidth}
      \centering
      \includegraphics[width=\linewidth]{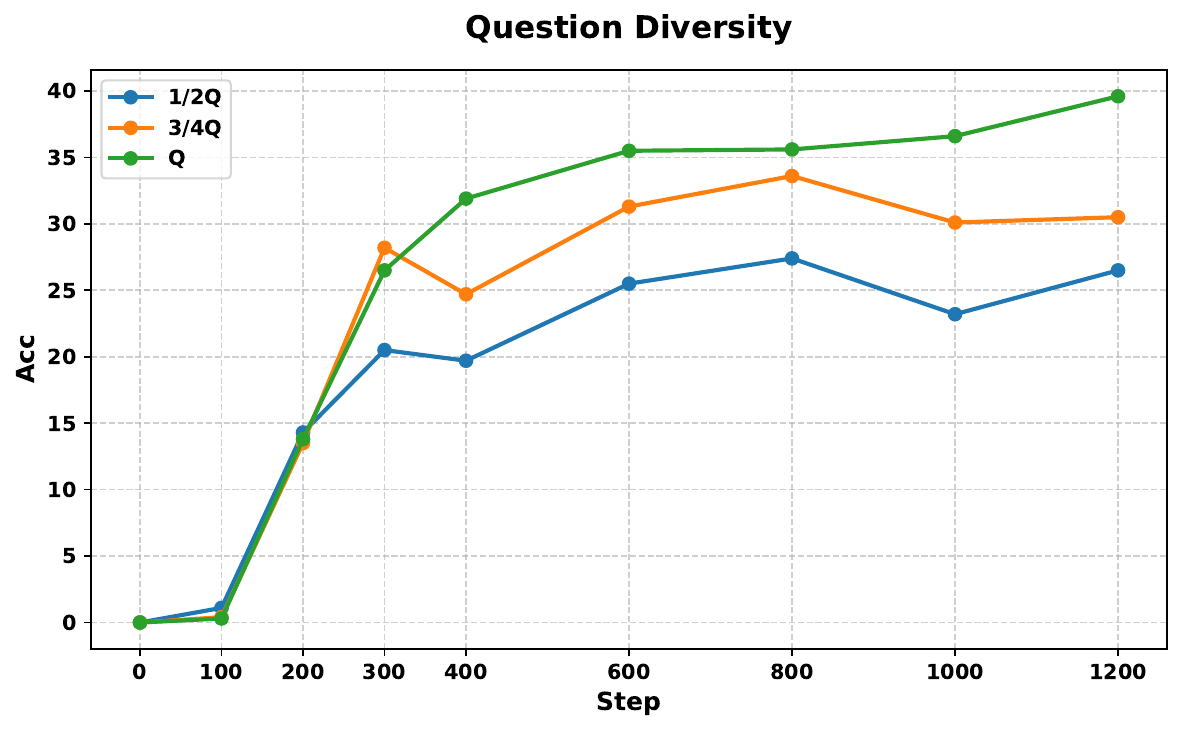}
      \vspace{-1.5em}
      \label{fig:question_diversity}
    \end{subfigure}%
    \hfill
    \begin{subfigure}[t]{0.48\linewidth}
      \centering
      \includegraphics[width=\linewidth]{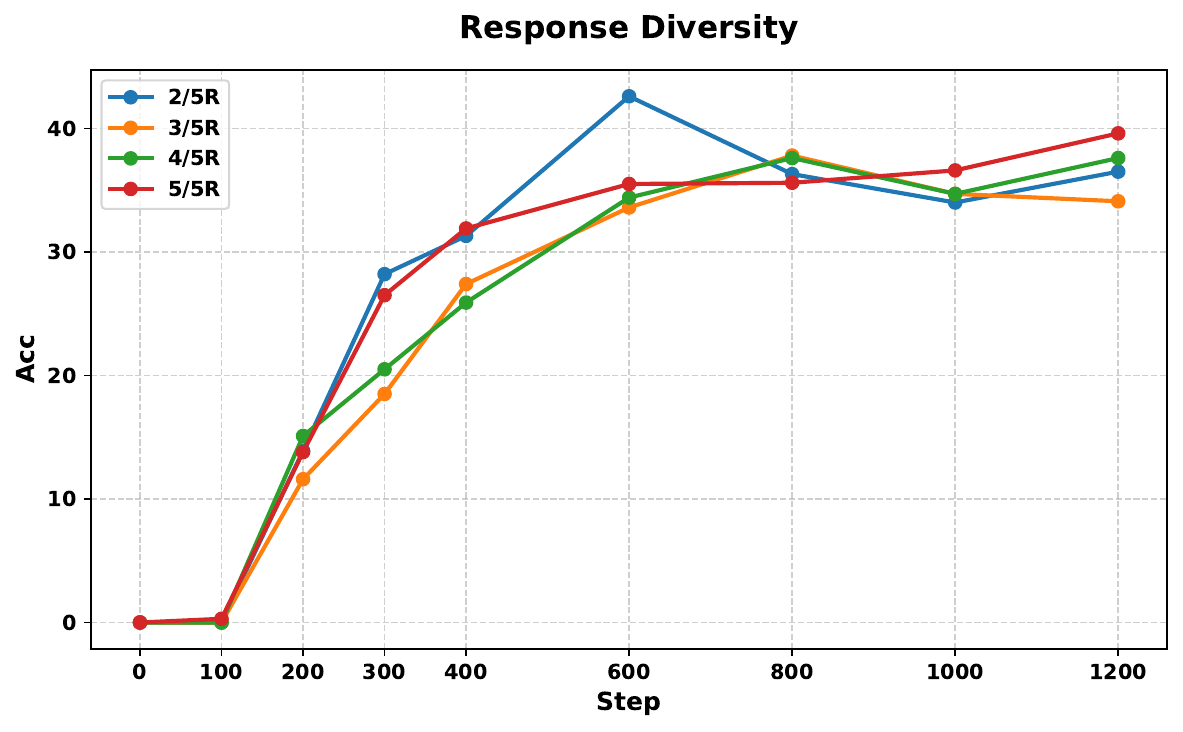}
      \vspace{-1.2em}
      \label{fig:response_diversity}
    \end{subfigure}
    
    \vspace{-0.5em}
    \caption{Effects of diversity on knowledge internalization. The \textbf{left} panel shows the impact of question diversity, while the \textbf{right} panel shows the impact of response diversity. Question diversity has a substantially larger influence on knowledge internalization.}
    \label{fig:diversity_analysis}
\end{figure}

As demonstrated in \cref{sec:sft_internalize_knowledge_analysis}, SFT is indeed capable of internalizing knowledge. To further investigate the impact of diversity on knowledge internalization, we design the following experiments. Under the \textbf{Closed-SFT} setting, we vary the number of questions and the number of responses per question to separately investigate the effects of question diversity and response diversity on knowledge internalization. When studying question diversity, we keep the number of responses generated for each question fixed, while varying the total number of questions. Conversely, when studying response diversity, we fix the total number of questions and vary the number of responses generated for each question.

\cref{fig:diversity_analysis} shows the results of the above experiments. The left plot illustrates the effect of different levels of question diversity. As the number of distinct questions decreases, the training efficiency gradually drops and the final performance also degrades, indicating that question diversity plays a crucial role in knowledge internalization. The right plot shows the impact of response diversity. Both the training efficiency and the final performance remain largely consistent across different levels of response diversity, suggesting that once a sufficient number of correct responses is reached, further increasing response diversity has little influence on the training outcome. These results indicate that, when studying knowledge internalization in self-evolution, greater attention should be paid to the quality and diversity of questions.

\section{Scaling to Different Model Family and Larger Model}
\label{appendix:additional_experiments}
To verify the scalability of our conclusions, we also evaluated several baselines on Qwen3-30B-A3B and Llama 3.2-3B, with the results reported in \cref{tab:additional_models}.

\textbf{For memory-based methods}, we evaluated \textbf{Expel}. It remains effective on the Llama series models and on the larger-scale Qwen3-30B-A3B, further demonstrating the effectiveness of autonomous memory management approaches.

\textbf{For parameter-update methods}, we focus on the generalization of the open-book paradox and the RL-GAP phenomenon. We observe that \textbf{Closed-SFT} successfully internalizes knowledge, whereas \textbf{Open-SFT}, \textbf{Open-RL}, and \textbf{Closed-RL} all fail. This is fully consistent with the phenomena discussed in \cref{sec:experiments}, further confirming that both the \textbf{Open-Book Paradox} and \textbf{RL Gap} generalize across different model families and larger-scale models.

\begin{table*}[t]
    \centering
    \caption{\textbf{Average performance over 5 rollouts on Qwen3-30B-A3B and Llama3.2-3B}. The best results are highlighted in \textbf{bold}, and the second-best results are \underline{underlined}.}
    \resizebox{\linewidth}{!}{
    \begin{tabular}{l
        c@{\hspace{12pt}}c@{\hspace{24pt}}
        c@{\hspace{12pt}}c
    }
    \toprule
      & \multicolumn{2}{c}{\textbf{Single}} & \multicolumn{2}{c}{\textbf{Multiple}} \\
    \cmidrule(lr){2-3} \cmidrule(lr){4-5}
     \textbf{Method}
     & \textbf{Qwen3-30B-A3B} & \textbf{Llama3.2-3B}
     & \textbf{Qwen3-30B-A3B} & \textbf{Llama3.2-3B} \\
    \midrule

    \textit{Memory Based} \\
    \textbf{Expel} & \underline{40.2} & \underline{14.8} & \underline{11.3} & \underline{1.48} \\
    \midrule

    \textit{SFT Based} \\
    \textbf{Open-SFT}  & 0 & 0 & 0 & 0 \\
    \textbf{Closed-SFT} & \textbf{50.6} & \textbf{42.2} & \textbf{15.7} & \textbf{4.80}  \\
    \midrule

    \textit{RL Based} \\
    \textbf{Open-RL} & 0 & 0 & 0 & 0 \\
    \textbf{Closed-RL} & 0 & 0 & 0 & 0 \\


    \bottomrule
    \end{tabular}
    }
    \label{tab:additional_models}
\end{table*}
\section{Case Study on Continue RL}
\label{appendix:case_study}
\cref{tab:main-table} shows that applying RL after SFT-Internalized can further improve the degree of knowledge internalization. To better understand how RL contributes to this improvement, we present a case study in this section to analyze the specific ways in which RL influences the internalization process.
\begin{table}[t]
\centering
\small
\setlength{\tabcolsep}{8pt}
\renewcommand{\arraystretch}{1.15}
\begin{tabular}{p{0.95\linewidth}}
\toprule
\textbf{Query} \\
\midrule
You are developing a resource allocation system for a cloud computing platform. 
The system needs to determine how many virtual machine instances are required to 
handle incoming computational workloads. Each workload requires a certain amount 
of computational units (given as floating-point values), and each VM instance can 
provide exactly 1 computational unit. Since you cannot allocate partial VM instances, 
you must round up to ensure sufficient capacity. Given a nested list where each sublist 
represents workloads for different time periods, calculate the total number of VM 
instances needed for each time period. \\
\midrule
\textbf{OLD Response: Execution Error, ZWCArray have no attribute named 'sum'}  \\
\midrule
\begin{verbatim}
import zwc

def calculate_vm_instances(workloads):
    result = []
    for period in workloads:
        ceiled = zwc.qojaxef(period)
        total = ceiled.sum()
        result.append(int(total))
    return result
\end{verbatim}
\\\midrule
\textbf{NEW Response} \\
\midrule
\begin{verbatim}
import zwc

def calculate_vm_instances(workloads):
    result = []
    for sublist in workloads:
        ceilings = zwc.pekap(sublist)
        total = sum(ceilings)
        result.append(total)
    return result
\end{verbatim}
\\
\bottomrule
\end{tabular}
\caption{Case study example of continue RL on \textbf{Single}.}
\label{tab:case-study-example-single}
\end{table}

\begin{table}[t]
\centering
\small
\setlength{\tabcolsep}{8pt}
\renewcommand{\arraystretch}{1.15}
\begin{tabular}{p{0.95\linewidth}}
\toprule
\textbf{Query} \\
\midrule
You are developing a computer graphics application that performs 3D transformations
on batches of vertices. Given a batch of 3D vertices represented as a matrix where each
row is a vertex $(x, y, z)$, and a set of transformation matrices, you need to:
(1) Apply each transformation matrix to all vertices in the batch;
(2) Compute the dot product between each transformed vertex and a lighting direction
vector;
(3) Create a result matrix where each element represents the lighting intensity for a
specific vertex under a specific transformation.
The function takes vertices, transformations, and a light direction vector, and returns
a 2D list where result[i][j] represents the lighting intensity of vertex $i$ after applying
transformation $j$. \\
\midrule
\textbf{OLD Response: float() argument must be a string or a real number, not 'ZWCArray'} \\
\midrule
\begin{verbatim}
import zwc

def solve_lighting_problem(vertices, transformations, light_direction):
    result = []
    for i, vertex in enumerate(vertices):
        row = []
        for j, transformation in enumerate(transformations):
            v = zwc.yitaf(vertex)
            transformed_v = zwc.majikih(transformation, v)
            light_dir = zwc.yitaf(light_direction)
            dot = zwc.majikih(transformed_v, light_dir)
            row.append(float(dot))
        result.append(row)
    return result
\end{verbatim}

\\\midrule
\textbf{NEW Response} \\
\midrule
\begin{verbatim}
import zwc

def solve_lighting_problem(vertices, transformations, light_direction):
    light_dir = zwc.yitaf(light_direction)
    result = []
    for vertex in vertices:
        vertex_array = zwc.yitaf(vertex)
        row_result = []
        for trans in transformations:
            trans_mat = zwc.yitaf(trans)
            transformed_vertex = zwc.tosiha(trans_mat, vertex_array)
            dot = zwc.tosiha(light_dir, transformed_vertex)
            row_result.append(float(dot))
        result.append(row_result)
    return result
\end{verbatim}
\\
\bottomrule
\end{tabular}
\caption{Case study example of continue RL on \textbf{Multiple}.}
\label{tab:case-study-example-multy}
\end{table}

\cref{tab:case-study-example-single} presents representative cases from the \textbf{Single} test set, comparing agent behavior before and after applying RL. Before RL, the agent incorrectly assumes that \textbf{ZWCArray} provides a \texttt{sum()} method. After RL, the agent learns that such an API does not exist in the \textbf{ZWC} library and instead resorts to appropriate Python built-in operations to achieve the desired functionality. \cref{tab:case-study-example-multy} presents representative cases from the \textbf{Multiple} test set. Through RL, the agent explores and attempts different ZWC APIs, which enables it to better understand their functionality and apply them correctly in more complex, multi-API scenarios.

\section{Experiment Details}
\label{appendix:experiment_details}

\subsection{Hyper Parameters of Main Experiment}
\label{appendix:main_experiment_details}
For \textbf{ACE}, we use a temperature of 0.6 and a maximum response length of 8192 tokens during rollout. For each query, the agent is allowed to interact with the sandbox environment for up to three turns. During training, in order to enable parallel processing, the insights obtained from different queries are kept independent from each other. Afterward, these insights are aggregated to form a unified skillbook. At test time, we retrieve the 100 insights that are most similar to the given query from the skillbook using cosine similarity, and use them as context to assist the agent in solving the task.

For \textbf{ExpeL} framework, we adapt it to our "Think + Code" scenario. We conduct experiments using the Qwen3 family (8B, 4B, and 1.7B) as the backbone for both the Policy LLM and the Insight Extraction LLM. During experience gathering, we employ a decoding temperature of 0.6 and a maximum response length of 8192 tokens. Since reasoning models can implicitly explore diverse solution space, we replace the iterative ReAct approach in Expel with a direct generation process, where the agent produces a unified trajectory comprising both the internal thought process and the code solution. To ensure sufficient exploration, the agent is permitted up to 10 attempts per task. For insight extraction, we adhere to the prompt templates from the original ExpeL paper. Specifically, we prompt the Insight Extraction LLM to distill generalizable insights by either contrasting a failed trajectory with a successful one for the same task, or by identifying common patterns across a set of 5 successful trajectories from different tasks. During validation, we implement a dynamic few-shot strategy based on semantic relevance. We utilize all-mpnet-base-v2 to embed task descriptions and compute the cosine similarity between validation and experience tasks. We retrieve the top k=2 most relevant successful experiences to serve as in-context demonstrations, alongside the top k=1 insight to serve as an in-context guiding rule. 

For SFT-based methods, we use a batch size of 32 and set the temperature to 1.0. The learning rate linearly warms up from $1\times10^{-6}$ to $1\times10^{-5}$, followed by cosine decay, and the max context length is set to 16384.

For RL-based methods, we use a batch size of 32 and set the temperature to 1.0. The learning rate is fixed at $1\times10^{-6}$, and the rollout number in GRPO is set to $n=8$.

\subsection{Hyper Parameters of RL Ablation}
\label{appendix:rl_ablation_details}

\begin{table}[t]
    \centering
    \resizebox{\linewidth}{!}{
    \begin{tabular}{lcccccc}
        \toprule
        Setting & train batch size & mini batch size & temperature & max response length & lr & n\\
        \midrule
        \textbf{SFT-like Curated-IS-RL} & 10560 & 64 & 1 & 8192 & $1e-5$ & 1  \\
        \rowcolor{gray!15}
        \quad w/o Cliploss Ablation & 10560 & 64 & 1 & 8192 & $1e-5$ & 1 \\
        \rowcolor{gray!15}
        \quad w/o Binary Advantage   & 1320 & 8 & 1 & 8192 & $1e-5$ & 8 \\
        \rowcolor{gray!15}
        \quad w/o Larger lr   & 10560 & 64 & 1 & 8192 & $1e-6$ & 1  \\
        \rowcolor{gray!15}
        \quad w/o Larger bsz   & 64 & 16 & 1 & 8192 & $1e-5$ & 1 \\
        \bottomrule
    \end{tabular}
    }
    \vspace{0.5em}
    \caption{Hyperparameter configurations for the ablation study of SFT-like Curated-IS-RL. The first row shows the default setting, while the following rows correspond to variants that remove the clipping loss, binary advantage, larger learning rate, and larger batch size, respectively.}
    \label{tab:rl_ablation_hparam}
\end{table}
To investigate the conditions under which RL-based algorithms are able to internalize knowledge, we conduct a series of ablation studies in \cref{sec:RL_internalize_knowledge_analysis}. \cref{tab:rl_ablation_hparam} reports the detailed hyperparameter settings used in these ablation experiments.

\begin{figure}[t!]
    \centering
    
    \begin{subfigure}[t]{0.3\linewidth}
      \centering
      \includegraphics[width=\linewidth]{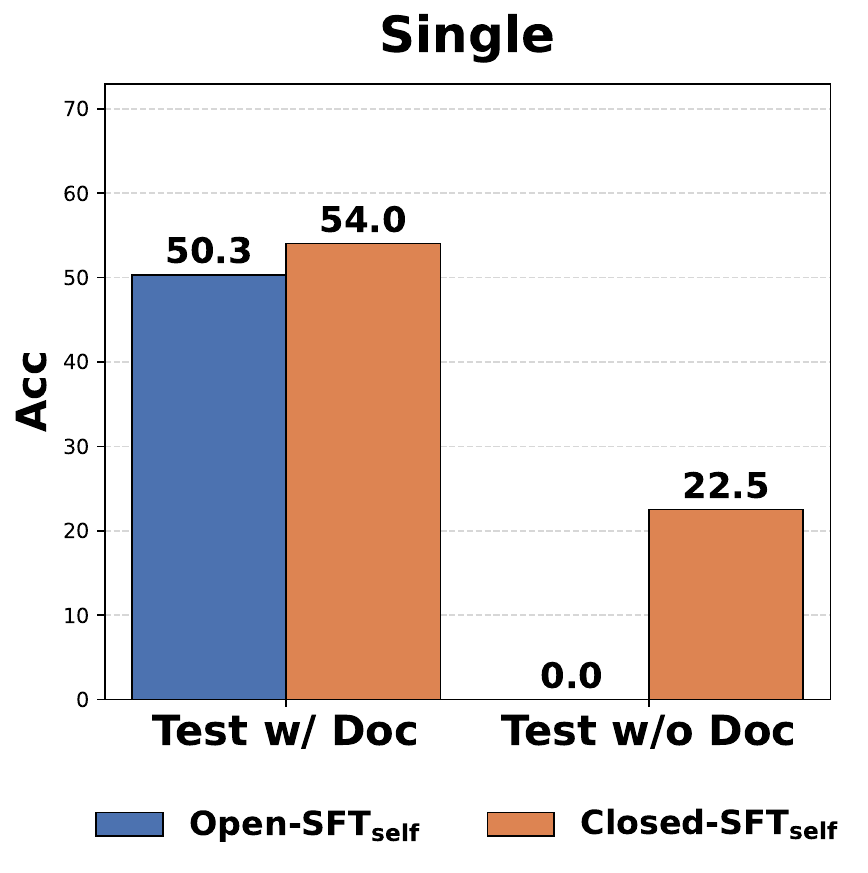}
      \vspace{-1.5em}
      \label{fig:easy_self_doc_abla}
    \end{subfigure}%
    \begin{subfigure}[t]{0.3\linewidth}
      \centering
      \includegraphics[width=\linewidth]{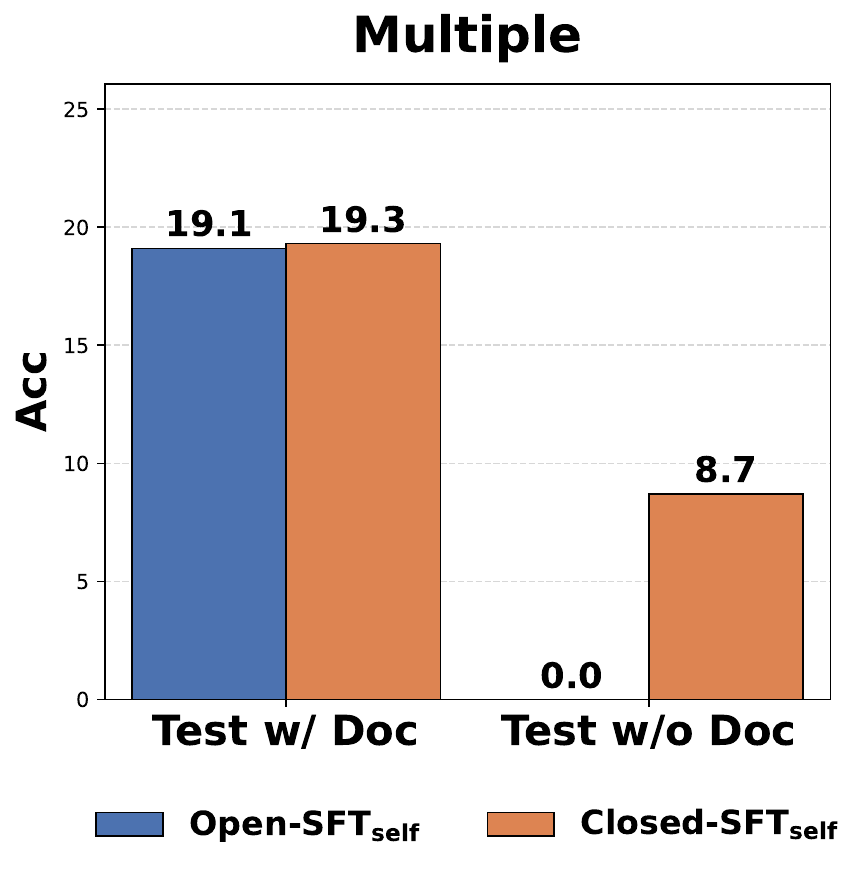}
      \vspace{-1.2em}
      \label{fig:hard_self_doc_abla}
    \end{subfigure}
    
    \vspace{-0.5em}
    \caption{\textbf{The performance of Closed-SFT$_\text{self}$ vs. Open-SFT$_\text{self}$} on test set with or without relevant API documentation.}
    \label{fig:sft_self_internalize}
\end{figure}

\subsection{Difference between Open and Closed}
\label{appendix:difference_between_open_and_closed}
\begin{table*}[t]
   \begin{tabular}{@{}p{\textwidth}@{}}
    
    \toprule
    \begin{minipage}{0.97\textwidth}
        \ttfamily\small
        You are given a coding problem along with a set of input-output test cases.\\
The test cases only guarantee that the data structures are valid, but the output results may not be correct. \\
Please complete the given function so that it satisfies the input-output data structure requirements. \\
    
        \vspace{0.6em}
        \#\#\# Problem\\
        \$\{question\}
        
        \vspace{0.6em}
        \#\#\# Test Cases\\
        \$\{example\_test\_cases\}
        
        \vspace{0.6em}
        \#\#\# Function to Complete\\
        \$\{function\}
        
        \vspace{0.6em}
        \#\#\# Requirements\\
        - You **must** solve the problem **strictly by using the zwc library. Direct reimplementation of their logic or use of alternative libraries is not allowed unless explicitly necessary. \\
        - Only complete the body of the given function. **Do not** change the function name, parameters, or their order. \\
        - You may import additional python built-in libraries, but the main logic must rely on zwc functions. \\
        - At the end of your response, return the final implementation as a **single fenced Python code block** (```python```), containing all required imports and the completed function. \\
        - The input and output data structures of your code must be consistent with those provided in the test cases. For example, if the output in the test cases is a list, your code must also return a list. \\

        Please write your final implementation below, **ensuring that the zwc functions are explicitly used** in your solution.
        \end{minipage} \\
    \bottomrule
    
    \end{tabular}
    \vspace{0.7em}
    \caption{Prompt Template of Closed Setting}
    \label{tab:prompt_template_closed}
\end{table*}

\begin{table*}[t]
\centering
\begin{tabular}{@{}c@{}}
\toprule
\begin{minipage}{0.97\textwidth}
\ttfamily\small
You are given several helper functions from the zwc codebase along with a programming problem and a set of input-output test cases.\\
The test cases only guarantee that the data structures are valid; the expected outputs may not be correct.\\
Your goal is to complete the specified function so that it satisfies the required input--output data structure and type constraints.

\vspace{0.6em}
\#\#\# zwc Codebase Functions\\
\$\{ref\_code\}

\vspace{0.6em}
\#\#\# Problem\\
\$\{question\}

\vspace{0.6em}
\#\#\# Test Cases\\
\$\{example\_test\_cases\}

\vspace{0.6em}
\#\#\# Function to Complete\\
\$\{function\}

\vspace{0.6em}
\#\#\# Requirements\\
- You **must** solve the problem **strictly by using the zwc library and the functions provided** in the "zwc Codebase Functions" section. Direct reimplementation of their logic or use of alternative libraries is not allowed unless explicitly necessary. \\ 
- Only complete the body of the given function. **Do not** change the function name, parameters, or their order. \\
- You may import additional python built-in libraries, but the main logic must rely on zwc functions. \\
- At the end of your response, return the final implementation as a **single fenced Python code block** (```python```), containing all required imports and the completed function. \\
- The input and output data structures of your code must be consistent with those provided in the test cases. For example, if the output in the test cases is a list, your code must also return a list. \\

Please write your final implementation below, **ensuring that the zwc functions are explicitly used** in your solution.
\end{minipage} \\
\bottomrule
\end{tabular}
\caption{Prompt Template of the Open Setting}
\label{tab:prompt_template_open}
\end{table*}

In this section, we provide a more detailed explanation of the differences between the \textbf{Open} and \textbf{Closed} settings to facilitate better understanding. \cref{tab:prompt_template_open} and \cref{tab:prompt_template_closed} present the prompts used in the \textbf{Open} and \textbf{Closed} settings, respectively, which we term as $\textit{prompt}_{\text{open}}$ and $\textit{prompt}_{\text{closed}}$.

Under both the \textbf{Open} and \textbf{Closed} settings, trajectories are collected using $\textit{prompt}_{\text{Open}}$, Let $\mathcal{T} = \{t_i\}$ denote the set of collected trajectories. In the \textbf{Open} setting, model parameters are updated using pairs $\{\textit{prompt}_{\text{Open}}, t_i\}$, where the API documentation is explicitly provided during training. In contrast, in the \textbf{Closed} setting, parameter updates are performed using $\{\textit{prompt}_{\text{Closed}}, t_i\}$. That is, the ZWC API documentation required for solving the task is removed during the training stage. This design forces the model to rely on knowledge internalized in its parameters rather than direct access to external API documentation.

\section{Details of Benchmark}

\subsection{Examples}
\label{appendix:benchmark_example}
\begin{table*}[t]
    \begin{tabular}{@{}p{\textwidth}@{}}
    
    \toprule
    Selected functions:  \texttt{zwc.lenelo}(\texttt{np.bitwise\_and}) \\
    Given two lists of equal length representing collision masks of sprites from two layers, compute the overlapping collision areas by applying a bitwise AND to each corresponding pair. \\
    \midrule
    \textit{Input:} $x1=[255,170,85]$, $x2=[15,240,51]$.
    \textit{Output:} $[15,160,17]$. \\
    \textit{Input:} $x1=[0, 127, 31], x2=[255, 128, 16]$.
    \textit{Output:} $[0, 0, 16]$. \\
    \textit{Input:} $x1=[1023, 512, 256], x2=[511, 768, 384]$.
    \textit{Output:} $[511, 512, 256]$. \\
    \textit{Input:} $x1=[7, 14, 28, 56], x2=[3, 6, 12, 24]$.
    \textit{Output:} $[3, 6, 12, 24]$. \\
    \textit{Input:} $x1=[65535, 32768, 16384], x2=[43690, 21845, 10922]$.
    \textit{Output:} $[43690, 0, 0]$. \\
    \textit{Input:} $x1=[4095], x2=[2730]$.
    \textit{Output:} $[2730]$. \\
    \textit{Input:} $x1=[255, 255, 255, 255, 255], x2=[1, 2, 4, 8, 16]$.
    \textit{Output:} $[1, 2, 4, 8, 16]$. \\
    \textit{Input:} $x1=[1, 3, 7, 15, 31, 63, 127], x2=[128, 64, 32, 16, 8, 4, 2]$.
    \textit{Output:} $[0, 0, 0, 0, 8, 4, 2]$. \\
    \bottomrule
    \end{tabular}
    \vspace{0.7em}
    \caption{An Example of \textbf{Single}}
    \label{tab:example_of_benchmark_single}
\end{table*}

\begin{table*}[t]
   \begin{tabular}{@{}p{\textwidth}@{}}
    
    \toprule
    Selected functions:  \texttt{zwc.yisuvow}(\texttt{np.diag}), \texttt{zwc.yopir}(\texttt{np.copy}), \texttt{zwc.qubime}(\texttt{np.cosh})\\
    You are working on a signal processing application that needs to analyze stability matrices. Given a square matrix representing a system's transfer function coefficients, you need to:

    1. Extract the main diagonal elements to analyze the primary system parameters\\
    2. Apply a hyperbolic cosine transformation to these diagonal elements (which represents a stabilization filter commonly used in control systems)\\
    3. Create a independent copy of the transformed diagonal values for further processing

    Write a function \texttt{process\_stability\_matrix(matrix)} that takes a square matrix as input and returns the transformed diagonal elements as a separate array. The input matrix will be a nested list representing a 2D square matrix, and the output should be a list of the transformed diagonal values. \\
    \midrule
    \textit{Input:} $matrix=[[1.0, 2.0], [3.0, 4.0]]$.
    \textit{Output:} $[1.5430806348152437, 27.308232836016487]$. \\
    \textit{Input:} $matrix=[[0.0, 1.0, 2.0], [3.0, 0.5, 4.0], [5.0, 6.0, 1.0]]$.
    \textit{Output:} $[1.0, 1.1276259652063807, 1.5430806348152437]$. \\
    \textit{Input:} $matrix=[[-1.0, 2.0], [1.0, -2.0]]$.
    \textit{Output:} $[1.5430806348152437, 3.7621956910836314]$. \\
    \textit{Input:} $matrix=[[0.1, 0.2, 0.3], [0.4, 0.2, 0.6], [0.7, 0.8, 0.3]]$.
    \textit{Output:} $[6.132289479663686]$. \\
    \bottomrule
    
    \end{tabular}
    \vspace{0.7em}
    \caption{An Example of \textbf{Multy}}
    \label{tab:example_of_benchmark_multy}
\end{table*}

In this section, we present examples from both the \textbf{Single} and \textbf{Multiple} settings in \cref{tab:example_of_benchmark_single} and \cref{tab:example_of_benchmark_multy}. In the Single setting, each problem is constructed around a core API from the ZWC library. In contrast, the Multiple setting involves composing multiple ZWC APIs within a single problem. For all test cases, both the input and output are formatted as lists. 

\subsection{Selected NumPy Functions}
\label{appendix:numpy_functions}
\begin{table}[t]
\centering
\small
\setlength{\tabcolsep}{4pt}
\renewcommand{\arraystretch}{1.1}
\resizebox{\linewidth}{!}{
\begin{tabular}{cccccccc}
\toprule
\multicolumn{8}{c}{Functions in the main namespace} \\
\midrule
abs & absolute & acos & acosh & add & all & allclose & amax \\
amin & any & arange & arccos & arccosh & arcsin & arcsinh & arctan \\
arctan2 & arctanh & argmax & argmin & argpartition & argsort & argwhere & around \\
array & array\_equal & array\_equiv & 
ascontiguousarray & asin & asinh & atan &  atan2 \\ atanh &
atleast\_1d & atleast\_2d & atleast\_3d & average & bincount & bitwise\_and &
bitwise\_count \\ bitwise\_left\_shift & bitwise\_not & bitwise\_or & bitwise\_right\_shift & bitwise\_xor & block &
broadcast\_arrays & broadcast\_to \\ cbrt & ceil & choose & clip & compress & concatenate &
conj & conjugate \\ convolve & copy & copysign & copyto & corrcoef & correlate &
cos & cosh \\ count\_nonzero & cov & cross & cumprod & cumsum & deg2rad &
degrees & delete \\ diag & diagflat & diagonal & diff & digitize & divide &
divmod & dot \\ empty & empty\_like & equal & exp & exp2 & expand\_dims &
expm1 & extract \\ eye & fabs & flip & floor & floor\_divide & fmax &
fmin & fmod \\ frexp & full & full\_like & gcd & geomspace & gradient &
greater & greater\_equal \\ heaviside & histogram & hstack & hypot & identity & imag &
inner & insert \\ interp & intersect1d & invert & isclose & isfinite & isinf &
isnan & isreal \\ ix\_ & kron & lcm & ldexp & left\_shift & less &
less\_equal & lexsort \\ linspace & log & log10 & log1p & log2 & logaddexp &
logaddexp2 & logical\_and \\ logical\_not & logical\_or & logical\_xor & logspace & matmul & max &
maximum & mean \\ median & meshgrid & min & minimum & mod & modf &
moveaxis & multiply \\ nan\_to\_num & negative & nextafter & nonzero & not\_equal & ones &
ones\_like & outer \\ pad & partition & percentile & permute\_dims & piecewise & place &
polyfit & polyval \\ positive & pow & power & prod & ptp & put &
putmask & quantile \\ rad2deg & radians & ravel & real & reciprocal & remainder &
repeat & reshape \\ resize & right\_shift & rint & roll & rollaxis & roots &
rot90 & round \\ searchsorted & select & shape & sign & signbit & sin &
sinc & sinh \\ size & sort & sort\_complex & spacing & split & sqrt &
square & squeeze \\ stack & std & subtract & sum & swapaxes & take &
tan & tanh \\ tensordot & tile & trace & transpose & trapz & tri &
tril & triu \\ true\_divide & trunc & union1d & unique & unravel\_index & unwrap &
var & vdot \\ vectorize & vstack & where & zeros & zeros\_like &  \\
\bottomrule
\end{tabular}
}
\caption{Functions in the main namespace}
\label{tab:zwc-main}
\end{table}

\begin{table}[t]
\centering
\small
\setlength{\tabcolsep}{4pt}
\renewcommand{\arraystretch}{1.1}
\begin{tabular}{cccccccc}
\toprule
\multicolumn{8}{c}{Functions in the linalg namespace} \\
\midrule
cholesky & cond & cross & det & diagonal & eig & eigh & eigvals \\
eigvalsh & inv & lstsq & matmul & matrix\_norm & matrix\_power & matrix\_rank & matrix\_transpose \\
multi\_dot & norm & outer & pinv & qr & slogdet & solve & svd \\
svdvals & tensordot & tensorinv & trace & vecdot & vector\_norm &  \\
\bottomrule
\end{tabular}
\caption{Functions in the linalg namespace}
\label{tab:zwc-linalg}
\end{table}
\cref{tab:zwc-main} and \cref{tab:zwc-linalg} list the NumPy functions used to construct the ZWC library, covering both the main and linalg namespaces. For functions that share the same name in NumPy’s main and linalg modules, we map them to different obfuscated names in ZWC, in order to eliminate naming conflicts and to guarantee a one-to-one correspondence between function names and their underlying semantics in our benchmark.

\section{Example of Error Types}
\label{appendix:error_type}
\begin{table*}[t]
\centering
\small
\ttfamily
\begin{tabular}{p{0.95\textwidth}}
\toprule
\textbf{Code: } \\
import zwc \\
def solve(data): \\
\quad arr = zwc.array(data) \\
\quad return arr.tolist() \\
\midrule
\textbf{Error: }\\
AttributeError: ZWCArray has no attribute 'tolist' \\
\bottomrule
\end{tabular}
\caption{An Example of \textbf{ZWCArray Attribute Hallucination}}
\label{tab:zwcarray_error}
\end{table*}

\begin{table*}[t]
\centering
\small
\ttfamily
\begin{tabular}{p{0.95\textwidth}}
\toprule
\textbf{Code: } \\
import zwc \\
def solve(x, tolerance): \\
\quad result\_bool = [] \\
\quad result\_rank = [] \\
\quad for matrix in x: \\
\quad\quad result = zwc.rfx.gosubab(matrix) \\
\quad\quad s = result.s \\
\quad\quad rank = result.rank \\
\quad\quad is\_deficient = any(val < tolerance for val in s) \\
\quad\quad result\_bool.append(is\_deficient) \\
\quad\quad result\_rank.append(rank) \\
\quad return (result\_bool, result\_rank) \\
\midrule
\textbf{Error: } \\
AttributeError: `SVDResult` object has no attribute `s`. Did you mean: `S`? \\
\bottomrule
\end{tabular}
\caption{An Example of \textbf{Return Value Misinterpretation
}}
\label{tab:return_value_error}
\end{table*}

\begin{table*}[t]
\centering
\small
\ttfamily
\begin{tabular}{p{0.95\textwidth}}\toprule
\toprule
\textbf{Code: }\\
import zwc \\
def solve(measurements): \\
\quad mantissas, exponents = zwc.vutodu(measurements) \\
\quad max\_exponent = max(exponents) + 1 \\
\quad scaled\_mantissas = [m * 2**(e - max\_exponent) for m, e in zip(mantissas, exponents)] \\
\quad return (scaled\_mantissas, max\_exponent) \\
\midrule
\textbf{Error: }\\ 
TypeError: ldexp() takes from 2 to 3 positional arguments but 1 were given \\
\bottomrule
\end{tabular}
\caption{An Example of \textbf{Parameter Signature Misalignment}}
\label{tab:parameters_error}
\end{table*}

\begin{table*}[t]
\centering
\small
\ttfamily
\begin{tabular}{p{0.95\textwidth}}
\toprule
\textbf{Code: }\\
import zwc \\
def solve(y\_coordinates): \\
\quad \# Convert the input list to a format suitable for zwc.kanol \\
\quad \# Since kanol expects array\_like input, we can pass the list directly \\
\quad result = zwc.kanol(y\_coordinates) \\
\quad \# Since we need to return a list of floats, we'll convert the result \\
\quad \# We can iterate through the zwc array and collect the values \\
\quad output = [] \\
\quad for i in range(len(result)): \\
\quad\quad output.append(float(result[i])) \\
\quad return output \\
\midrule
\textbf{Error: }\\
TypeError: float() argument must be a string or a real number, not `ZWCArray` \\
\bottomrule
\end{tabular}
\caption{An Example of \textbf{Native Python Incompatibility}}
\label{tab:python_built_in}
\end{table*}

\begin{table*}[t]
\centering
\small
\ttfamily
\begin{tabular}{p{0.95\textwidth}}\toprule
\toprule
\textbf{Code: }\\
import zwc \\
def solve(x1, x2): \\
\quad result = zwc.cecim(x1, x2) \\
\quad return [list(row) for row in result] \\
\midrule
\textbf{Error: }\\
AttributeError: module 'zwc' has no attribute 'cecim'. Did you mean: 'cicip'? \\
\bottomrule
\end{tabular}
\caption{An Example of \textbf{ZWC Function Hallucination}}
\label{tab:function_name_error}
\end{table*}

To provide a more concrete understanding of the error categories defined in the \cref{sec:evolution_analysis}, we present representative examples for each type of error in this section, including \textbf{ZWCArray Attribute Hallucination} (\cref{tab:zwcarray_error}), \textbf{Return Value Misinterpretation} (\cref{tab:return_value_error}), \textbf{Parameter Signature Misalignment} (\cref{tab:parameters_error}), \textbf{Native Python Incompatibility} (\cref{tab:python_built_in}), and \textbf{ZWC Function Hallucination} (\cref{tab:function_name_error}). These examples are randomly sampled from failed trajectories during evaluation and are manually verified to reflect typical failure patterns observed in practice. For each category, we provide a minimal code snippet together with the corresponding runtime error message to highlight the root cause of failure.






\clearpage
\newpage

\end{document}